\documentclass[runningheads]{llncs}
\usepackage{bbding}
\usepackage{lmodern}
\usepackage{siunitx}
\usepackage{booktabs}
\usepackage{etoolbox}
\usepackage{bm}
\usepackage{graphicx}
\usepackage{tabularx}
\usepackage{subcaption}
\usepackage{xcolor}
\usepackage{soul}
\usepackage{mwe}
\usepackage{caption}
\usepackage{float}
\usepackage{enumitem} 
\usepackage{makecell} 
\usepackage{breqn}
\usepackage{hyperref}

\begin{document}

\title{Privileged Anatomical and Protocol Discrimination in Trackerless 3D Ultrasound Reconstruction}
\titlerunning{Privileged Information in Trackerless 3D Ultrasound Reconstruction}

\author{
Qi Li\inst{1} \and
Ziyi Shen\inst{1} \and
Qian Li\inst{1,2} \and
Dean C. Barratt\inst{1} \and
Thomas Dowrick\inst{1} \and
Matthew J. Clarkson\inst{1} \and
Tom Vercauteren\inst{3} \and
Yipeng Hu\inst{1}} 

\authorrunning{Q. Li et al.}
\institute{Centre for Medical Image Computing, Wellcome/EPSRC Centre for Interventional and Surgical Sciences, Department of Medical Physics and Biomedical Engineering, University College London, London, U.K.\\
\and
State Key Laboratory of Robotics and System, Harbin Institute of Technology, Harbin, China \and
School of Biomedical Engineering \& Imaging Sciences, King’s College London, London, U.K.
}

\maketitle    
\begin{abstract}
Three-dimensional (3D) freehand ultrasound (US) reconstruction without using any additional external tracking device has seen recent advances with deep neural networks (DNNs). In this paper, we first investigated two identified contributing factors of the learned inter-frame correlation that enable the DNN-based reconstruction: anatomy and protocol.
We propose to incorporate the ability to represent these two factors - readily available during training - as the privileged information to improve existing DNN-based methods. This is implemented in a new multi-task method, where the anatomical and protocol discrimination are used as auxiliary tasks. We further develop a differentiable network architecture to optimise the branching location of these auxiliary tasks, which controls the ratio between shared and task-specific network parameters, for maximising the benefits from the two auxiliary tasks.
Experimental results, on a dataset with 38 forearms of 19 volunteers acquired with 6 different scanning protocols, show that 1) both anatomical and protocol variances are enabling factors for DNN-based US reconstruction; 2) learning how to discriminate different subjects (anatomical variance) and predefined types of scanning paths (protocol variance) both significantly improve frame prediction accuracy, volume reconstruction overlap, accumulated tracking error and final drift, using the proposed algorithm.

\end{abstract}
\section{Introduction}\label{Introduction}
3D ultrasound (US) reconstruction is a promising technique both for diagnostics and image guidance, such as image fusion with other image modalities \cite{lindseth2003multimodal}, registration with preoperative data during surgery \cite{lang2012multi}, volume visualisation and measurement \cite{guo2022ultrasound}.
Currently, most clinically applied 3D reconstruction of 2D freehand US imaging use spatial tracking devices, such as optical, electromagnetic or mechanical positioning \cite{mozaffari2017freehand}. Research for reducing dependency on external devices in such 3D US reconstruction has been motivated for its portability, accessibility and low-cost. 
Previous non-learning-based methods utilised the speckle correlation between US frames \cite{chang20033,lang2012multi}. 
In recent learning-based approaches, convolutional neural networks and their variants have been proposed to use two adjacent frames as network input \cite{miura2020localizing,prevost20183d,prevost2017deep}, for predicting spatial transformation between them. More generally, sequential modelling methods, e.g. using recurrent neural network \cite{li2022trackerless,luo2021self,luo2022deep,miura2021pose} and transformer \cite{ning2022spatial}, have also been tested with the same goal of localising relative positions between two or more US frames. 

With the promising results from deep learning, one might question what was learned in these data-driven approaches for predicting inter-frame transformations. Specifically, in addition to speckle patterns (which holds within a limited spatial scale), what are other factors that generated correlation between US frames, such that their relative locations can be inferred?

We first hypothesized that two factors, common anatomical characteristics between subjects and predefined scanning protocols, are responsible for such predictability in this application. A variance-reduction study is presented in Sec. \ref{atm_ptc_variance_exp} to demonstrate that sufficient anatomical and protocol variance in training data is indeed required for the reconstruction.

In this work, we propose to encode the anatomical and protocol patterns using two classification tasks, discriminating between subjects and between types of scanning paths, respectively. We then investigate methods to train these tasks together with the main reconstruction task to improve the performance of the main task.

In addition to US images as network input, previous studies have also integrated additional information or signals, such as optical flow \cite{xie2021image} between US frames and acceleration, orientation, angular rates from inertial measurement units (IMU)~\cite{luo2022deep,mikaeili2022trajectory}. Optical flow was derived from image sequence itself, while the IMU-measured signals are required at both training and inference. Different to these applications, the proposed discrimination task labels, subject and protocol indices, are in general available during training, but not required at inference, thus known as privileged information \cite{vapnik2009new}, further discussed in Sec. \ref{sec:method_preliminary}.

In summary, our contributions include 1) a multi-task learning approach to formulate two factors in freehand US as privileged information, for improving reconstruction accuracy; 2) a mixture model formulation for optimisable branching locations for auxiliary tasks, implemented with a differentiable network and a gradient-based bi-level optimisation; 3) extensive experimental results to quantify the improved performance due to the privileged tasks; and 4) open-source code and data \footnote{\url{https://github.com/ucl-candi/freehand}} for public access and reproducibility.

\section{Method}

\subsection{Preliminaries: Privileged Auxiliary Tasks and Shared Parameters}\label{sec:method_preliminary}
Assume a main task $f_{\theta}(y|x)$ that predicts ${y}$ with input data $x$ (here, using a $\theta$-parameterised neural network), which is optimised alongside $\mathcal{J}$ auxiliary tasks $f_{\theta^{s}_j, \theta^{a}_j}(y^{a}_j|x)$, $j=1,...,\mathcal{J}$, where $\theta^{s}_j$ are shared parameters (with the main task), and $\theta^{a}_j$ are task-specific parameters, both for predicting $y^{a}_j$. Therefore, $\theta^{s}_j \subseteq \theta$ and $\theta^{a}_j \cap \theta = \emptyset$. When the goal is to predict $y$ from $x$, the supervision and prediction of auxiliary task are only required during training. The main task benefits from this privileged information, which is not required at inference.

Such multi-task learning incorporates the privileged information but may suffer from absolute negative transfer \cite{wu2020understanding} - here, when the additional auxiliary tasks negatively impact the main task performance, and/or relative negative transfer - one auxiliary task reduces the main-task-improving potentials from the other auxiliary task(s). One approach for reducing negative transfer, or optimising the transferability, is adjusting the ratio between the shared $\theta^{s}_j$ and task-specific $\theta^{a}_j$ parameters \cite{newell2019feature}. As no parameters are shared (i.e. $\theta^{s}_j = \emptyset$), no negative transfer is possible (although any benefit from this auxiliary task also diminishes). 

\begin{figure}[H]
        \centering
        \begin{subfigure}{\textwidth}
            \centering
            \includegraphics[width=0.6\textwidth]{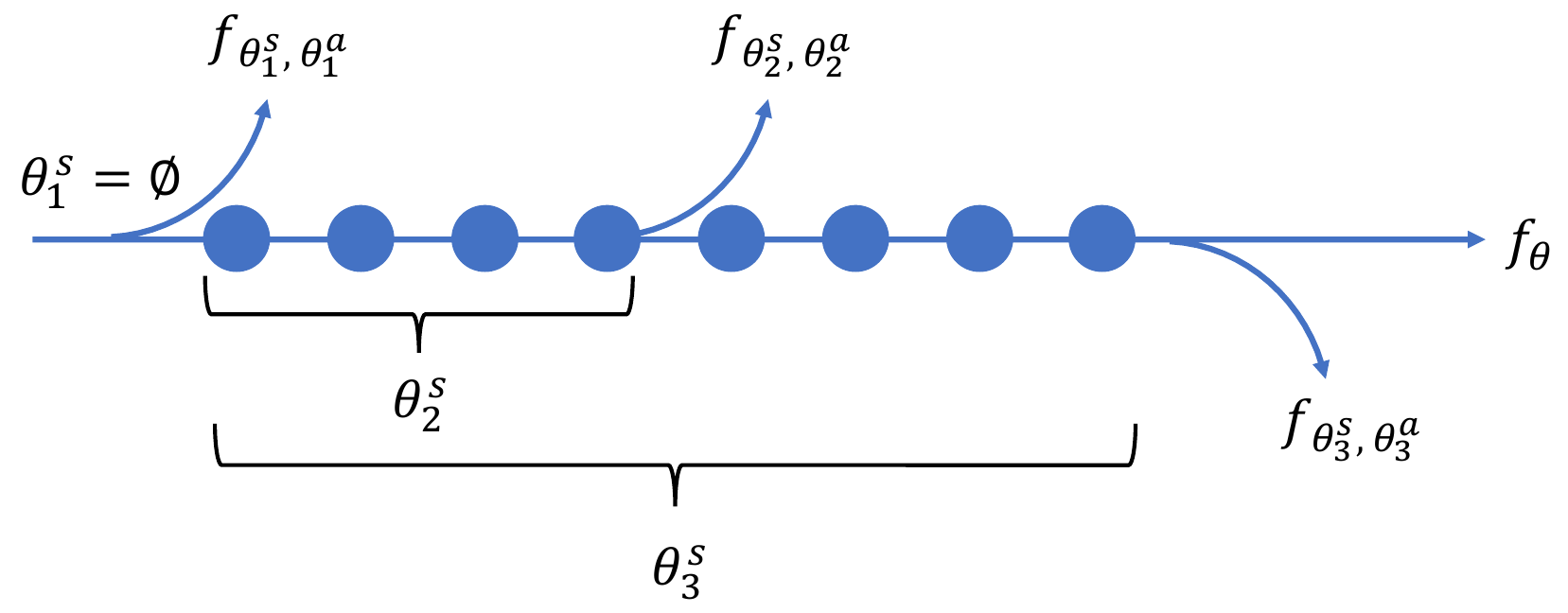}
            \caption[]%
            {{\small }}    
            \label{MTL-1}
        \end{subfigure}
        \hfill
        \begin{subfigure}{\textwidth}
            \centering
            \includegraphics[width=0.75\textwidth]{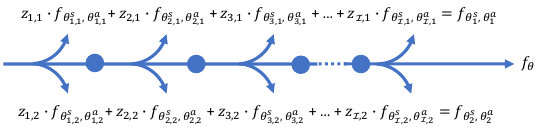}
            \caption[]%
            {{\small }}    
            \label{MTL-2}
        \end{subfigure}
        \caption[ ]
        {\small Example of network architectures with (a) various possible branching locations with different shared parameters and (b) two auxiliary tasks, each modelled as a mixture of $\mathcal{I}$ candidate tasks.} 
        \label{mtl-paradigm-fig}
\end{figure}

Assume a binary task descriptor $z_j$ (extended to a one-hot vector in Sec.~\ref{mtl-paradigm}) indicating the use of the $j^{th}$ auxiliary task. The main task is thus conditioned on the use of the auxiliary task $f_{\theta,\theta^{a}_{j}}(y|x,z_j)$. 
Importantly, where to place the task descriptor $z_j$ determines which and how many network parameters $\theta^{s}_j$ are shared. As in Fig.~\ref{mtl-paradigm-fig} (a), the closer the task descriptor is to the input, i.e. early branching, the less shared parameters.

\subsection{Example Auxiliary Tasks: Anatomical and Protocol Discrimination}\label{atm_ptc_variance}
Each auxiliary task is trained with a cross-entropy loss, between predicted class probability vectors $y^{a}_{j}$ and ground-truth targets $t^{a}_{j}$ in one-hot vectors.
\begin{equation} \label{eq:loss_aux}
\mathcal{L}^{CE}_j = -\sum_{N_j}(t^{a}_{j} \cdot log[f_{\theta^{s}_{j}, \theta^{a}_{j}}(y^{a}_{j}|x)])
\end{equation}

The two tasks, minimising $\mathcal{L}^{CE}_{j=1}$ and $\mathcal{L}^{CE}_{j=2}$, classify different training subjects and types of scanning protocols, with $N_{j=1}$ and $N_{j=2}$ number of classes, respectively.
An underlying assumption is that the anatomical and protocol variance can impact the 3D US reconstruction performance. This assumption is tested by quantifying the reconstruction accuracy changes, as number of subjects and/or protocol types are reduced. Results are reported in Sec. \ref{atm_ptc_variance_exp}.   

\subsection{Parameterised Task Descriptor Locations}\label{mtl-paradigm}
Assume $\mathcal{I}$ locations in the main task network, at which a task descriptor $z_{i,j}$ can be conditioned, where $i=1,...,\mathcal{I}$ for $j^{th}$ auxiliary task. For a single main task in this case, each auxiliary task can branch out from these $\mathcal{I}$ locations, as illustrated in Fig.~\ref{mtl-paradigm-fig} (b).
The task descriptor thus represents the probability of branching location, with additional constraints $z_{i,j} \in [0,1]$ and $\sum_iz_{i,j}=1$. 

With the task descriptor $z_{i,j}$, the $j^{th}$ auxiliary task is parameterised by a mixture model of $\mathcal{I}$ candidate tasks $f_{\theta^{s}_{i,j}, \theta^{a}_{i,j}}(y^{a}_{i,j}|x)$, each performed by one branch, where the additional subscript $i$ is the candidate task index.

\begin{equation}
f_{\theta, \theta^{a}_{1,j}, ..., \theta^{a}_{\mathcal{I},j}} (y^{a}_j|x,z_j)=\sum_{i=1}^{\mathcal{I}}[z_{i,j} \cdot f_{\theta^{s}_{i,j}, \theta^{a}_{i,j}}(y^{a}_{i,j}|x)]
\end{equation}
where $z_j=[z_{i,j}]^\top_{i=1,...,\mathcal{I}}$ is the task descriptors for all the auxiliary tasks, with shared parameters $\theta^{s}_j=[\theta^{s}_{i,j}]^\top_{i=1,...,\mathcal{I}}$ and task-specific parameters $\theta^{a}_j=[\theta^{a}_{i,j}]^\top_{i=1,...,\mathcal{I}}$. For $j^{th}$ auxiliary task, the final prediction is therefore the matrix product of all candidate (branch) predictions $y^{a}_j=[y^{a}_{i,j}]^\top_{i=1,...,\mathcal{I}}$ (a $N_j \times \mathcal{I}$ matrix) and the location weights $z_j$ (a $\mathcal{I} \times 1$ vector, generated by a softmax function). Multiplicative task conditioning is also known as gating. The loss defined in Eq.~\ref{eq:loss_aux} remains.

This formulation allows all candidate branches trained together without a  predefined architecture decision. We show in Sec.~\ref{Meta-training} that the task descriptor may be considered as a hyperparameter and optimised using a gradient-based bi-level optimisation algorithm, for efficient inference using a single branch. 

\subsection{The Main Task and Evaluation}\label{Evaluation_metrics}       
We adopt the approach in our previous work \cite{li2022trackerless} for the main task, reconstructing a 3D US scan by sequentially predicting inter-frame transformations with inputs of frame sequences. The same reconstruction loss is used, but is now conditioned on the two auxiliary tasks, denoted as $\mathcal{L}^{Rec}(t,f_{\theta,\theta^{a}_{1},\theta^{a}_{2}}(y|x,z_1,z_2))$
where the network output is the six parameters for rigid spatial transformation. It is important to clarify that this proposed loss function also utilises a multi-task learning for predicting transformations between nearby frames, but different to and independent of that used in this work. Unless specified otherwise, the methodology and the implementation of the main task network, based on EfficientNet (b1) \cite{tan2019efficientnet}, remain the same, with further details in the original publication~\cite{li2022trackerless}.

To test the generalisation and reconstruction performance of the proposed method, four evaluation metrics are used. For each frame, \textit{frame prediction accuracy} ($\epsilon_{frame}$) is used to evaluate the generalisation of the method, denoting the Euclidean distance between the ground-truth and prediction- transformed four corner points of each frame. The scan reconstruction performance is quantified by an \textit{accumulated tracking error} ($\epsilon_{acc.}$), indicating the averaged point distance on all frame pixels, a \textit{volume reconstruction overlap} ($\epsilon_{dice}$), denoting the overlap of all pixels between the ground-truth and prediction volume, and a \textit{final drift} ($\epsilon_{drift}$), denoting the \textit{frame prediction accuracy} of the last frame in a scan.          

\subsection{Bi-level Optimisation of Task Descriptor}\label{Meta-training}

Given a loss for the main task $\mathcal{L}^{Rec}(t,f_{\theta,\theta^{a}_{1},\theta^{a}_{2}}(y|x,z_1,z_2))$ and those for the two auxiliary tasks defined in Eq.~\ref{eq:loss_aux}, the overall loss function thus is:
\begin{equation}\label{eq:loss_overall}
\mathcal{L}(\theta,\theta^{a}_{1},\theta^{a}_{2}, z_1, z_2|D) = \mathcal{L}^{Rec}(\theta|x,t) + \sum_{j=1}^2 \mathcal{L}^{CE}_j(\theta,\theta^{a}_{j},z_j|x, t^{a}_1, t^{a}_2)
\end{equation}
where the loss is rearranged as a function of relevant network parameters $\{\theta,\theta^{a}_{1},\theta^{a}_{2}\}$ and task descriptors $\{z_1, z_2\}$, with observed data $D=\{x,t, t^{a}_1, t^{a}_2\}$. Given a training data set $\mathcal{D}_{train}$ and a validation data set $\mathcal{D}_{val}$, the empirical losses are $\mathcal{L}_{train}(\cdot)=\mathbf{E}_{D \in \mathcal{D}_{train}}[\mathcal{L}(\cdot|D)]$ and $\mathcal{L}_{val}(\cdot)=\mathbf{E}_{D \in \mathcal{D}_{val}}[\mathcal{L}(\cdot|D)]$, respectively. Optimising the task descriptors subject to optimised network parameters leads to the following meta-learning task with a bi-level optimisation problem:
\begin{multline} \label{bilevel}
\hat{z_1}, \hat{z_2} = \arg\mathop{\min}_{z_1, z_2} \mathcal{L}_{val} (\hat{\theta},\hat{\theta}^{a}_{1},\hat{\theta}^{a}_{2}; z_1, z_2) \\
\textrm{s.t.}~~ \hat{\theta},\hat{\theta}^{a}_{1},\hat{\theta}^{a}_{2} = \arg\mathop{\min}_{\theta,\theta^{a}_{1},\theta^{a}_{2}} \mathcal{L}_{train} (\theta,\theta^{a}_{1},\theta^{a}_{2}; z_1, z_2)
\end{multline} 

Since the $\frac{\partial \mathcal{L}_{val}}{\partial z_j}$ can be estimated, the task descriptors can be optimised by gradient-based updates alternating between minimising the two empirical loss functions \cite{rajeswaran2019meta}. The numerical algorithm is summarised in Algorithm 1. The three tasks are weighed equally in Eq.~\ref{eq:loss_overall} as the task descriptor hyperparameters make explicit optimising or predefining these weights redundant.

\begin{table}
\begin{minipage}{\linewidth}
\begin{tabular}{l}
\hline
\textbf{Algorithm 1:} The bi-level optimisation algorithm.\\
\hline
1. Initialise task descriptors $\{z_1, z_2\}$ for two auxiliary tasks.\\
2. Optimise the task descriptors using meta-learning algorithm in a differentiable network:\\
$\,\, \, \, \, \, \,$\textbf{while} \emph{not converged} \textbf{do}\\
$\, \, \, \, \, \,$$\, \, \, \, \, \,$   (1) Update network parameters \{$\theta,\theta^{a}_{1},\theta^{a}_{2}$\} for the main task and all auxiliary tasks \\ 
$\, \, \, \, \, \,$$\, \, \, \, \, \,$ $\, \, \, \, \,\, \,\,$ by descending $\nabla_{\theta,\theta^{a}_{1},\theta^{a}_{2}} \mathcal{L}_{train} (\theta,\theta^{a}_{1},\theta^{a}_{2}; z_1, z_2)$\\
 $\, \, \, \, \, \,$$\, \, \, \, \, \,$   (2) Update task descriptors \{$z_1, z_2$\} \\
$\, \, \, \, \, \,$ $\, \, \, \, \, \,$ $\, \, \, \, \, \,\,\,$by descending $\nabla_{z_1, z_2} \mathcal{L}_{val} ((\theta,\theta^{a}_{1},\theta^{a}_{2})-\xi \nabla_{\theta,\theta^{a}_{1},\theta^{a}_{2}} \mathcal{L}_{train} (\theta,\theta^{a}_{1},\theta^{a}_{2}; z_1, z_2); z_1, z_2)$.\\

3. Finalise the network architecture using the optimised task descriptors \{$z_1, z_2$\}.\\
\hline
\end{tabular}
\footnotetext{In this work, the first-order approximation was used when optimising task descriptors with $\xi=0$, i.e., $\nabla_{z_1, z_2} \mathcal{L}_{val} (\theta,\theta^{a}_{1},\theta^{a}_{2}; z_1, z_2)$ \cite{liu2018darts}.}
\end{minipage}
\end{table}
\noindent

\section{Experiments and Results}
\subsection{Dataset and Network Settings}\label{Data_acquisition}
The US data used in this study was from our previously published data set in \cite{li2022trackerless}, acquired at 20 frame per second (fps) by an Ultrasonix machine (BK, Europe) with a curvilinear probe (4DC7-3/40, 6 MHz, 9 cm depth and a median level of speckle reduction) and an NDI Polaris Vicra (Northern Digital Inc., Canada) tracker.
All US images with a size of 480×640 pixels, after spatial and temporal calibration, acquired from 38 forearms of 19 volunteers were used in this study, more than 40,000 US frames in total. For each forearm, three predefined scanning protocols (straight line shape, `C' shape and `S' shape, as in Fig.~\ref{atm_ptc_var} (a)), in a distal-to-proximal direction, with the US probe perpendicular of and parallel to the forearm, were acquired, resulting in 6 different protocols and 228 scans in total. The US scans have a various number of frames, from 36 to 430 frames, equivalent to a probe travel distance of between 100 and 200 mm. 
The data was split into train, validation and test sets by a ratio of 3:1:1 on a scan level.

With the EfficientNet (b1) for the main reconstruction task~\cite{li2022trackerless}, nine locations were used for candidate branches, with each being a single fully-connected classification layer, denoted as Branches 1-9 and Branch 1 being closest to network input. The nine candidate predictions are weighted by the softmax-generated task predictor, as described in Sec. \ref{mtl-paradigm}, to form each final auxiliary task prediction. The two auxiliary tasks resulted in total of 18 branches. 

A minibatch of 32, an Adam optimizer were used for model training, with a learning rate of $10^{-4}$ (tested among $\{10^{-3}, 10^{-4}, 10^{-5}\}$) and the input sequence length being $M=\{100, 140\}$ (tested among $M=\{49, 75, 100, 140\}$), selected based on the validation set performance. Each model was trained for at least 15,000 epochs until convergence, for up to  4 days, on Ubuntu 18.04.6 LTS with a single NVIDIA Quadro P5000 GPU card. The model with the best validation set performance was selected to evaluate test set performance. Other hyperparameters were found relatively insensitive to model performance and configured empirically based on validation performance. 

\subsection{The Effect of Anatomical and Protocol Variances}\label{atm_ptc_variance_exp} 

\begin{figure}[H]
 \centering
\includegraphics[width=\textwidth]{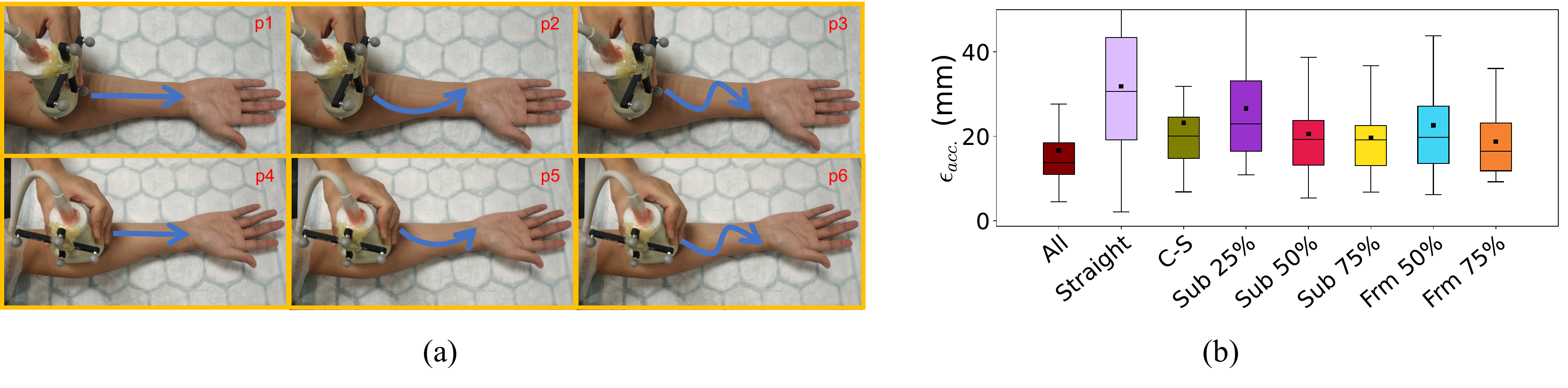}
\caption{Illustration of different US acquisition protocols and their imapct to reconstruction accuracy. (a) Protocols of straight line shape (p1, p4), `C' shape (p2, p5) and `S' shape (p3, p6) with US probe perpendicular of and parallel to the forearm, (b) $\epsilon_{acc.}$ changes due to reduced anatomical and protocol variance.} \label{atm_ptc_var}
\end{figure}

The impact on performance of the main task was tested by training models with various anatomical and protocol variance in the training data: 1) all training data (All); 2) straight only (Straight); 3) C-shape and S-shape (C-S); 4) 25\% subjects in training data (Sub 25\%); 5) 50\% training subjects (Sub 50\%); 6) 75\% training subjects (Sub 75\%); 7) 50\% of frames in a scan (Frm 50\%) and 8) 75\% of frames (Frm 75\%), and tested on the same test set. 
Fig.~\ref{atm_ptc_var} (b) plotted $\epsilon_{acc.}$ with $M=20$ as an example of the reconstruction performance using different training sets. The other models and metrics yielded a consistent trend, which saw $\epsilon_{acc.}$ increased with both reduced anatomical and protocol variances. It indicates both are factors impacting the main reconstruction task performance.  

\subsection{Ablation Studies and Comparison}

In our experiments, two types protocol discrimination tasks are considered, six-class classification (as in Fig. \ref{atm_ptc_var}) and three-class classification (combining the perpendicular and parallel scans for the same scan shapes). Each is combined with the 38-class classification task (38 training subjects) as the anatomical discrimination task. Results from different $M={100,140}$ are also included, together with those from the main task network without any branches (no-branch).

\begin{figure}[H]
 \centering
\includegraphics[width=\textwidth]{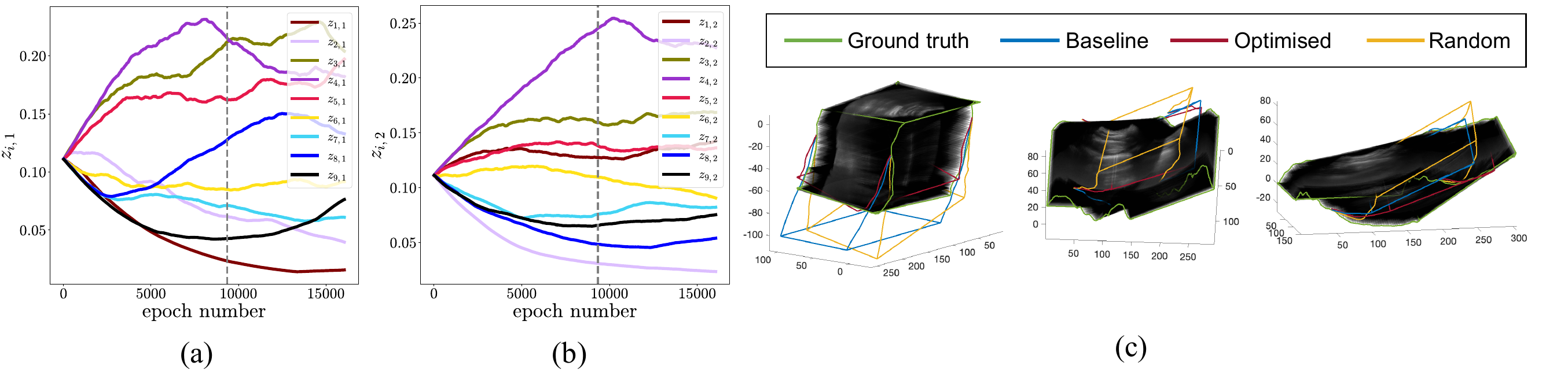}
\caption{The trend of task descriptor for anatomy (a) and protocol (b), $M=100$, and the reconstruction performance (c). The epoch indicating best performance of the model on validation set is denoted by a gray dotted line. 
Scans with various scanning path are reconstructed using no-branch, optimised branches, and random branches strategies, $M=100$.} \label{rec}
\end{figure}

The optimised $z_{i,j}$ values versus the training epochs are plotted in Fig.~\ref{rec} (a) and (b), where the optimum branch was selected by the maximum $z_{i,j}$ value, at the epoch.
Table.~\ref{proposed_method_two_branching_best} summarised the reconstruction performance of the proposed method, using the optimised branches (their indices are denoted with asterisks) for the two auxiliary tasks. Comparing with the no-branch models, the improved performances from the proposed methods can be seen, for both $M=100$ and $M=140$, regardless of the three- or six-class were used as protocol discrimination tasks. For example, at $M=100$, $\epsilon_{drift}$ was lowered from 14.52 to 6.56 mm, using the proposed methods with optimised Branches 4 and 4, for protocol and anatomical discrimination tasks, with $p$-value = 0.013 (unpaired t-test, at a significance level $\alpha=0.05$). Statistical significance was found in performance improvement using all four evaluation metrics. 

Although no previous work utilise these discrimination tasks as privileged information for assisting this application, two baseline models were implemented as alternatives to the main task network. We have re-implemented the proposed method using one of the first proposed approaches for this application \cite{prevost2017deep} as the main task network, with two adjacent frames as input and outputting the transformation between them. In addition, the same network architecture using more (10) input frames were also tested, denoted as `\cite{prevost2017deep}-10'. These results are also summarised in Table.~\ref{proposed_method_two_branching_best}, with and without optimised branches. However, we would like to emphasize that the reported inferior results from these compared baselines need to be interpreted with caution, as they were neither designed for nor tuned with incorporating these auxiliary tasks used in this study. They are included for completeness and reference values.

\begin{table}
\begin{center}
\caption{Mean and standard deviation of four metrics of the proposed method and no-branch method.}\label{proposed_method_two_branching_best}%
\tiny
\begin{tabular*}{\textwidth}{@{\extracolsep{\fill}}ccccccc@{\extracolsep{\fill}}}
\hline
$M$ &\makecell{Num. of \\protocols} & \makecell{Branch index \\(protocol/anatomy)}& $\epsilon_{frame}$& $\epsilon_{acc.}$   & $\epsilon_{dice}$ &  $\epsilon_{drift}$ \\
\hline

    100 &n/a & No-branch &  $0.18\pm0.05$ &  $7.03\pm3.97$ &  $0.84\pm0.08$&   $14.52\pm10.51$ \\     
    100 &       3 &     Branches $5^*/4^*$ &   $0.19\pm0.06$ &  $3.92\pm3.50$ &   $0.73\pm0.21$&   $7.06\pm7.30$ \\ 
    100  &    6   &   Branches $4^*/4^*$ &   $0.17\pm0.08$   &$3.80\pm3.97$ &  $0.76\pm0.24$&    $6.56\pm7.53$ \\ 
    140 &n/a& No-branch  & $0.14\pm0.05$ &  $3.68\pm3.10$ &   $0.62\pm0.28$ &  $7.30\pm7.40$ \\
    140 &       3  &     Branches $9^*/4^*$ &   $0.15\pm0.08$   &$3.36\pm3.26$ &   $\bm{0.94\pm0.00}$&   $\bm{6.20\pm6.31}$ \\ 
    140  &      6 & Branches $5^*/7^*$ &  $\bm{0.13\pm0.05}$  &$\bm{2.90\pm2.10}$ &    $0.89\pm0.00$&   $6.53\pm5.98$\\

    \cite{prevost2017deep} &  n/a  & No-branch& $0.59\pm0.28$ &  $29.03\pm9.15$ &  $0.43\pm0.32$ & $35.67\pm11.20$ \\ 
    \cite{prevost2017deep} &     3  &    Branches $1^*/9^*$  &   $0.70\pm0.50$   &$32.71\pm18.10$ &   $0.60\pm0.22$ & $59.53\pm36.87$ \\
    \cite{prevost2017deep}  &  6       &    Branches $4^*/9^*$  &   $0.68\pm0.46$   &$30.29\pm17.44$ &   $0.67\pm0.16$ & $55.05\pm32.69$ \\   
       
    \cite{prevost2017deep}-10& n/a &  No-branch&$0.38\pm0.21$ &  $17.60\pm9.77$ & $0.67\pm0.22$ &  $22.64\pm12.47$ \\ 
    \cite{prevost2017deep}-10    &    3 &     Branches $4^*/4^*$ &  $0.42\pm0.30$   &$19.37\pm10.63$ &   $0.50\pm0.30$ & $26.32\pm13.33$ \\  
    \cite{prevost2017deep}-10 & 6 &        Branches $9^*/4^*$ &  $0.43\pm0.38$   &$21.72\pm12.74$ &   $0.50\pm0.25$ & $29.64\pm16.29$ \\

\hline
\end{tabular*}
\end{center}
\end{table}

\section{Conclusion and Discussion}

This work demonstrated the impact of anatomical and protocol variance towards the 3D reconstruction of trackerless freehand US and formulated two respective discrimination tasks for taking advantage these privileged information during training. Using the proposed algorithm, substantially improved reconstruction performance was achieved, which may indicate a promising new direction for improving the potentials of this application for clinical adoption. Future work includes testing clinical applications with specific challenges, such as those without predefined protocol classes (where a clustering task may be used instead), and comparison with approaches such as gradient surgery \cite{yu2020gradient}, which may need adaptation for a single main task.

\subsubsection{Declarations.}
This work was supported by the EPSRC [EP/T029404/1], a Royal Academy of Engineering / Medtronic Research Chair [RCSRF1819\textbackslash7\textbackslash734] \makebox[\textwidth][s]{(TV), Wellcome/EPSRC  Centre  for Interventional  and  Surgical  Sciences} \newline[203145Z/16/Z], and the International Alliance for Cancer Early Detection, an alliance between Cancer Research UK [C28070/A30912; C73666/A31378], Canary Center at Stanford University, the University of Cambridge, OHSU Knight Cancer Institute, University College London and the University of Manchester. TV is co-founder and shareholder of Hypervision Surgical. Qi Li was supported by the University College London Overseas and Graduate Research Scholarships. For the purpose of open access, the authors have applied a CC BY public copyright licence to any Author Accepted Manuscript version arising from this submission. This study was performed in accordance with the ethical standards in the 1964 Declaration of Helsinki and its later amendments or comparable ethical standards. Approval was granted by the Ethics Committee of local institution (UCL Department of Medical Physics and Biomedical Engineering) on $20^{th}$ Jan. 2023 [24055/001].


\bibliographystyle{splncs04}
\bibliography{paper}

\end{document}